\documentclass[a4paper]{article}

\usepackage{INTERSPEECH2020}
\usepackage{multirow}
\title{LSTM-LM with Long-Term History for First-Pass Decoding in Conversational Speech Recognition}
\name{Xie Chen, Sarangarajan Parthasarathy, William Gale, Shuangyu Chang, Michael Zeng}
\address{
 Microsoft, USA}
\email{\{xieche, sarangp, William.Gale, Shawn.Chang, nzeng\}@microsoft.com}
\begin{document}
% \ninept
\maketitle

\begin{abstract}
LSTM language models (LSTM-LMs) have been proven to be powerful
and yielded significant performance improvements over 
count based $n$-gram LMs in modern speech recognition systems.
Due to its infinite history states and computational load, most previous 
studies focus on applying LSTM-LMs in the second-pass for rescoring purpose. Recent work shows that
it is feasible and computationally affordable
to adopt the LSTM-LMs in the first-pass decoding within a 
dynamic (or tree based) decoder framework.
In this work, the LSTM-LM is composed with a WFST decoder on-the-fly for the first-pass decoding. Furthermore,
motivated by the long-term history nature of LSTM-LMs, the use of context beyond the current utterance is explored for the first-pass decoding in conversational speech recognition. The context information is captured by the hidden states of LSTM-LMs across utterance and can be used to guide the first-pass search effectively. The experimental results in our internal meeting transcription system show 
that significant performance improvements can be obtained by incorporating the contextual information with LSTM-LMs in the first-pass decoding, compared to applying the contextual information in the second-pass rescoring.
\end{abstract}


\section{Introduction}
\label{sec:intro}
LSTM-LMs \cite{sundermeyer2012interspeech} are attracting more and more research interests for speech recognition systems and significant performance improvements are reported over traditional $n$-gram LMs in a range of tasks \cite{sundermeyer2015taslp}. 
LSTM-LMs are able to capture the long-term history through the recurrent connection and continues history representation, which could address the short-term history and sparsity issues existed in $n$-gram LMs.
However, compared to the use of count based $n$-gram LMs for speech recognition, there are several potential drawbacks which prevent the use of LSTM-LMs. The probability of LSTM-LMs is slow to compute. In $n$-gram LMs, the probability computation is very efficient and mainly consists of table look-up operation. While for the LSTM-LM, the matrix multiplication and non-linear function are computationally heavy, especially for tasks with large vocabulary. In order to mitigate this issue, self-normalization techniques \cite{mnih2013nips, chen2015ICASSP, shi2014icassp} were proposed to approximate the probability without normalization in the output layer. Furthermore, the $n$-gram LM probability only depends on the previous $n$-1 words. The finite history and back-off feature in $n$-gram LMs are very suitable for HMM based decoder infrastructure where different paths can be recombined when their $n$-gram LM histories are identical. In contrast, the LSTM-LMs require the complete history for probability computation. As a result, theoretically, recombination can only happen for paths with the same history. Various approximations were introduced for the path recombination for lattice rescoring of LSTM-LMs, e.g. $n$-gram approximation \cite{liu2014icassp, sundermeyer2014interspeech, xu2018icassp}. Given these concerns, in most previous studies, LSTM-LMs were mainly used for the second-pass rescoring, e.g. n-best or lattice rescoring.  Two recent works \cite{jorge2019lstmlm, beck2019arxiv} demonstrate that it is feasible and computationally affordable to apply LSTM-LMs in the first-pass decoding and comparable performances are achieved to use LSTM-LMs in the first-pass decoding compared to second-pass rescoring.

LSTM-LMs are powerful to capture long-term information and are able to improve both perplexity and word error rate (WER) in the state-of-the-art ASR systems. In most situations, speech recognition is operated on utterance level and LSTM-LMs are applied to capture history information from the beginning of the utterance. In conversational speech, history information beyond the current utterance may contain various information such as grammar, use of word, and topic, which are expected to be helpful to improve language model and ASR systems. 
 There are several previous work investigating the long-term information to improve speech recognition \cite{chelba2017arxiv, tuske2018interspeech, xiong2018EMNLP, irie2019asru}. The use of long-term history information could further improve the performance of language model and speech recognition. However, all of these works applied LMs carrying cross utterance information in the second pass rescoring, on top of the n-best or lattice generated in the first-pass decoding. 
 In traditional HMM based ASR system, which is still the main-stream system in practice to date, $n$-gram LMs are usually applied for the first-pass decoding. $n$-gram LMs only consider the previous $n-1$ words and it lacks the ability of capturing long-term history beyond current utterance. The traditional approach is to use $n$-gram LMs for the first-pass decoding and n-best list (or lattice) generation. In order to incorporate LMs with long-term history, more complicated language models including LSTM-LMs, MaxEnt LMs can be applied in the second-pass rescoring. However, the candidates produced in the first-pass decoding do not contain any context information and maybe not optimal. The search space is limited by the $n$-gram LMs. The use of LMs with context information in the second-pass rescoring can only re-rank candidates among these sub-optimal candidates, which may cause mismatches. 
 
 Motivated by the long-term history nature of LSTM-LMs and recent advance on LSTM-LMs for the first-pass decoding, in this paper, we explore the use of long-term history for LSTM-LMs in the first-pass decoding to fully utilize the context information in conversational speech, where context is crucial for people to understand and communicate with each other.
The experimental results in a real meeting transcription system demonstrate that the use of long-term context can improve the performance of speech recognition system consistently and significantly compared to the use of LSTM-LMs with long-term context in the second pass rescoring.

This paper is organized as follows. Section \ref{relatedwork} briefly reviewed the related work for the use of context for speech recognition and LSTM-LMs for the first-pass decoding. Section \ref{sec:lstminwfst} introduces the implementation of LSTM-LMs for the first-pass decoding in a WFST decoder, followed by the details on the incorporation of the context information in the first-pass decoder with LSTM-LMs in Section \ref{sec:context}.
The experimental results are presented in Section \ref{sec:exps}. Finally, the paper is concluded and future work is discussed in Section \ref{sec:conclusion}.

\section{Related Work}
In this section,  previous works on using LSTM-LMs for the first-pass decoding and long-term history for improving language model are briefly reviewed.
\label{relatedwork}
\subsection{LSTM-LM for First-Pass Decoding}
In the past several years, LSTM-LMs \cite{mikolov2010interspeech, sundermeyer2012interspeech} have been proven to yield significant performance improvement over traditional $n$-gram LMs in speech recognition. However, due to their computational cost and long-term history dependency, it was believed not practical for LSTM-LMs in the first-pass decoding. Instead, LSTM-LMs were adopted for the second-pass rescoring extensively in most previous works. Recent progress \cite{jorge2019lstmlm, beck2019arxiv} demonstrates the potential of using LSTM-LMs for the first-pass decoding on the tree based decoder. In order to speedup the decoding, it is necessary to cache history vectors of LSTM-LMs to avoid repeated computation for LSTM-LMs probabilities with same histories. Self-normalized probability is also applied to reduce computation in the output layer with large vocabulary. There are also several interesting common observations from the results reported in \cite{jorge2019lstmlm, beck2019arxiv}. One is that the $n$-gram LM is used for look-ahead and the LSTM-LM probability was added for the language model score update. In addition, both of these two works compared the use of $n$-gram approximation \cite{liu2014icassp, sundermeyer2014interspeech} for the history recombination with different $n$. The experiment results showed that the best performance was achieved when $n$ is set to infinity, where histories are merged only if they are exactly the same.

\subsection{Improving LM with Long-Term History}
There have been many previous works on the use of context to improve language models for speech recognition. In \cite{kuhn1990pami}, the cache-based language model was trained on the most recent utterances and interpolated with the background LM. This adapted language model was then applied for following utterances. In this way, the adapted $n$-gram LM is able to implicitly utilize the context information to boost the probabilities of the $n$-grams appeared before. However, it requires updating the language model frequently with the change of history and it is difficult to collect sufficient context utterances for robust LM adaptation in practice.
\cite{gildea1999ecsct, gildea1999ecsct, watanabe20csl, bellegarda2000ieee} investigated the use of topic information to build improved $n$-gram LMs for first-pass decoding.
In \cite{mikolov2012slt, chen2015interspeech}, the topic information was modelled in RNNLMs as additional features and used for rescoring. \cite{xiong2018acl} build a session-level LSTM-LM to capture the session-level information. The context dependent LSTM-LMs were applied in the second-pass decoding. They relies on the $n$-best (or lattice) candidates generated by $n$-gram LMs, which don't have context information. In \cite{chelba2017arxiv, tuske2018interspeech}, the LSTM based $n$-gram LM was constructed and applied for language modeling. Long-term history with crossing utterance, up to 40 history words, was investigated for second pass rescoring.
\cite{kim2019arxiv_v1, kim2019arxiv_v2} is more relevant to what we investigated in this paper, where the context information was applied explicitly for speech recognition in an end-to-end speech recognition system for conversational speech. In this work, we mainly focus on the traditional HMM-DNN based hybrid system since it is still the mainstream ASR system used in production.

\section{First-Pass Decoding with LSTM-LMs in WFST Decoder}
\label{sec:lstminwfst}
In previous work \cite{jorge2019lstmlm, beck2019arxiv}, LSTM-LMs were added dynamically in the tree based decoder, also known
as dynamic decoder.
Nowadays, static decoder based on WFST \cite{mohri2002csl} are becoming increasingly popular as it is faster and easier to optimize compared to dynamic decode. In this paper, LSTM-LMs were applied for the first-pass decoding by using WFST decoder.

\subsection{LSTM-LMs for WFST Decoder}
 In WFST based decoder, various knowledge sources, including acoustic model, pronunciation lexicon and $n$-gram language models, can be presented as finite state transducer and then composited into a static graph, which is usually known as $HCLG$ in literature \cite{mohri2002csl}. However, it is difficult to represent LSTM-LMs as FST because the history states of LSTM-LMs are infinite. Hence, the LSTM-LM probabilities have to be added dynamically during decoding, and the LSTM-LM probability will be interpolated with the $n$-gram LM probability stored in the WFST decoder.
However, during the construction of the $HCLG$ graph in the WFST decoder, the weight pushing operation \cite{mohri2002csl} pushes the LM score forward and the scores stored in the arcs are not the correct LM scores. For the interpolation between $n$-gram and LSTM LMs, it works fine for log-linear interpolation since the scores are added in log domain \footnote{note the log-linear interpolated probability is not normalized.}. However, it is not feasible for linear interpolation where the $n$-gram LM score in probability domain is required. In order to make it more flexible and able to support linear interpolation, the $HCLG$ graph is split into two FST graphs. The first graph adopts a pruned language model $G'$ to build $HCLG'$ and a delta grammar graph is then built for $G-G'$. In this way, the composition can be expressed as 
\begin{equation}
    HCLG' \circ (G-G')
\end{equation}
During decoding, these two FST are composited on-the-fly and the final language model scores are the same as the original $HCLG$ graph.
By using this on-the-fly composition, the underlying grammar defined by LSTM-LMs $G_{lstm}$ can be added dynamically through interpolating with the delta grammar $G-G'$. The composition supporting LSTM-LMs can be written as below,
\begin{equation}
    HCLG' \circ Intplt(G-G', G_{lstm}-G')
\end{equation}
where $Intplt$ denotes any valid interpolation functions, e.g. linear interpolation or log-linear interpolation. In this paper, we choose log-linear interpolation with equal weight for simplicity.

\subsection{Speed Optimization}
Another potential issue for the use of LSTM-LMs in the first-pass decoding is the efficiency. Similar to previous work \cite{jorge2019lstmlm, beck2019arxiv}, it is important to use unnormalized LSTM-LM probability during decoding and cache history vectors to avoid redundant computation. In this paper, Noise contrastive estimation \cite{gutmann2010icais} was used for training the self-normalized model. In order to predict the word $w_i$ giving history $w_{0}^{i-1}$ in LSTM-LMs, the history $w_{0}^{i-2}$ is represented using a continues history vector $\vec{h}_{i-2}$. In addition to caching the history vectors $\vec{h}_{i-2}$, the output of last hidden layer $\vec{s}_{i-1}$ and the LSTM-LM probability of the complete history and the current word $P(w_0^{i-1}, w_i)$ are also cached. When the LSTM-LM probability given history $w_0^{i-1}$ and word $w_i$ needs to be computed, the probability cache $P(w_0^{i-1}, w_i)$ is checked first if it is already existed. Otherwise, the final hidden layer output cache for $\vec{s}_{i-1}$ is then used to check whether the last hidden output has been computed. If so, we only need to compute for the output layer, which is fast due to the use of unnormalized probability. Lastly, the caching for the LSTM-LM vectors $\vec{h}_{i-2}$ will be fetched and the input word is set to $w_{i-1}$ for LSTM-LM forward to calculate the probability. All of these computations can be organized as batch mode as described in \cite{beck2019arxiv}.
In this paper, the LSTM-LM was trained using Pytorch \cite{paszke2017automatic}. LibTorch was used to export the Pytorch model into script mode, which can be called in C++ codes.

\section{LSTM-LMs with Long-term History for Speech Recognition}
\label{sec:context}
In most of the HMM based speech recognition systems, speech recognition is operated on utterance level after segmentation and the $n$-gram LM state is initialized with sentence start $<$s$>$. As discussed in Section \ref{relatedwork}, there are some previous works trying to adjust the $n$-gram LM used in the first-pass decoding according to the history, such as the cache based \cite{kuhn1990pami} and topic model based LMs \cite{iyer1996icslp}.  In this way, the LM may contain the context information implicitly by using the most recent words or embedding the topic information. In contrast, in this work, the context is incorporated explicitly in LSTM-LMs for the first-pass decoding.

In this paper, we mainly investigate the effect of context for conversational speech, where the contextual information is expected to play an important role for communication and understanding. Ideally, the LSTM-LMs should be trained using corpus consist of paragraphs with ordered utterances. However, in practice, most of the LM training corpus consists of utterances, instead of paragraphs. The order of utterance and paragraph information are not available. As a result, the LSTM-LMs trained on utterance level are adopted in order to leverage the large amount of LM training data. In spite of the potential mismatch between training and evaluation, LSTM-LMs are expect to be capable of capturing the context information across utterance due to their long-term history characters.  There is another potential due to the mismatch between the utterance level training and paragraph level evaluation. The sentence boundary $<$s$>$ only appears in the beginning of sentence during the LSTM-LM training, while it will appear in the middle of the sequence when the probability prediction is across utterances. The existence of the last $<$s$>$ in context may affect the prediction of first several words. In order to mitigate the effect, we also investigate the scenario without the last sentence boundary $<$s$>$ in the context \footnote{we only remove the last $<$s$>$ in the context and the sentence boundary $<$s$>$ in earlier utterances are kept, which are crucial for performance.}. The implementation to use context in LSTM-LMs is straightforward. When no context is used for the first-pass decoding, the LSTM-LM history vector is initialized to be all 0. When the context is available, the history vector of the LSTM-LM is initialized with the contextual utterances. In real-time speech recognition, the reference context is usually not available. The LSTM-LM history vector of the hypothesis from the previous utterance is then used as initial hidden vector for decoding.

It is worth noting that LSTM-LMs with context can be applied in the second-pass rescoring as well. While the beam search will probably prune the path in the first-pass decoding using $n$-gram LMs without context information. When LSTM-LMs with contextual information is applied in the first-pass decoding, it would be helpful to guide the search from the beginning of utterance. It is important to realize the search space with and without contextual information are different.

\section{Experiments}
\label{sec:exps}
\subsection{Experimental Setup}
In this paper, the experiments were conducted on an internal meeting transcription system, which is a typical scenario for conversational speech and where context is expected to be important. The acoustic model was trained on 6,4000 hours of audio using Layer Trajectory BLSTM \cite{eric2019interspeech} with sequence training.  A word list consisting of 250K words was chosen as vocabulary. The back-off 5-gram LM and LSTM-LM were trained on about 2.5 billion words. The LSTM-LM used in this work consists of 1 hidden layer with 1024 nodes. The $n$-gram and LSTM LMs are combined with log-linear interpolation using equal weight in the following experiments. When LSTM-LMs are applied for the first-pass decoding, there is no recombination for histories with different complete history.
It is worth noting that the LSTM-LM is trained on utterance level by resetting the history vector to 0 at the beginning of each utterance as most of our training corpus did not contain utterance order information.
Three meetings were selected as test set for performance evaluation, which consists of about 3 hours of speech. The average utterance length for the test set is 16 words. It is not rare for the overlap speech in the real meeting scenario. The overlap utterances are discarded for simplicity although it might affect the continuity of the context.
\subsection{Experimental Results}
Table \ref{tab:werbaseline} shows that the baseline WER results with LSTM-LM for the first-pass decoding and second-pass rescoring. For the second pass rescoring, $n$-best rescoring ($n$=100) was used. No context information was used for both $5$-gram and LSTM LMs. As expected, the LSTM-LM yielded significant performance improvement over the $5$-gram LM. The use of the LSTM-LM for the first-pass decoding gave comparable WER compared to the second-pass rescoring.
\begin{table}[htbp]
    \centering
    \caption{WER results of the meeting transcription system with LSTM-LM in the first-pass decoding and second-pass rescoring (100-nbest)}
    \begin{tabular}{|l||c|c|}
    \hline
       LM  &  \#pass    &   WER   \\
    \hline\hline
    5glm      & $1^{st}$     &   17.96     \\
    \quad+LSTM      & $1^{st}$     &   16.20     \\
    \quad+LSTM      & $2^{nd}$     &   16.26    \\
    \hline
    \end{tabular}
    \label{tab:werbaseline}
\end{table}

The next experiment is designed to investigate the effect of context for the LSTM-LM probability prediction.
Table \ref{tab:pplhist} demonstrates the PPL results of LSTM-LMs with different lengths of history context. The history utterances are used to initialize the hidden vectors of LSTM-LMs to predict the current utterance. The reference history was used for the PPL computation. Interestingly, the PPLs can be improved consistently with the increase of context. In this experiments, the reference history contains the $<$s$>$ symbols. The PPL plateaus when  the length of context is longer than 2 utterances. One potential explanation is that the mismatch between training and evaluation compensates the effect of longer context. The last line in Table \ref{tab:pplhist} gives the PPL result by shuffling the history utterances. In this case, the shuffled history is still relevant in content since they are from the same conversation, but the order information is missing. The shuffled complete history gives a PPL of 93.1, which is only 4 points lower than the PPL without context. The PPL improvement is about one fourth of the improvement when the ordered utterances are presented. It indicates that, compared to the content, the ordering information in the history contributes more for word prediction in following utterances.

\begin{table}[htbp]
    \centering
    \caption{PPL results of 5-gram and LSTM LMs with different numbers of history utterances (the average utterance length is 16 words).  $+\infty$ denotes that the complete history is used.} 
    \begin{tabular}{|c||c|c|c|}
    \hline
       LM     &  \#hist utterance &   PPL   \\
    \hline\hline
    5glm        & 0         &    146.9      \\
    \hline
    \multirow{6}{*}{LSTM}      
                & 0         &   96.9     \\
                & 1         &   84.3          \\
                & 2         &   82.4          \\
                & 4         &   81.4          \\
                & $+\infty$ &   82.1      \\
                & $+\infty$ (shuf) & 93.1   \\
    \hline
    \end{tabular}
    \label{tab:pplhist}
 \end{table}

Table \ref{tab:werlstmlonghist} shows the WER results for the first-pass decoding using LSTM-LMs with context information. According to the experimental results, the use of context in LSTM-LMs consistently improved the WER performance. The use of one previous history improved the WER from 16.20\% to 15.67\% (with reference history and w/o $<$s$>$). The complete reference history further reduce the WER to 15.51\%, about 4.3\% relative WER improvement. The WER improvement becomes marginal when there are more than 2 history utterances, which is consistent to the PPL experiment reported in Table \ref{tab:pplhist}. 
The statistical significance test was also carried out to compare WER with and without history context. It reveals that the WER performance differences are significant. The performances with context from reference and hypothesis were compared in Table \ref{tab:werlstmlonghist}. The reference history consistently outperformed the hypothesis history from recognition, the WER gap between reference and hypothesis history are negligible, less than 1\% relatively.    Table \ref{tab:werlstmlonghist} also reports the WER comparison with and without the last sentence boundary $<$s$>$ in context. 
According to the experimental results shown in Table \ref{tab:werlstmlonghist}, the absence of $<$s$>$ for the latest history utterance helps to improve the WER performance slightly but consistently.

\begin{table}[htbp]
    \centering
    \caption{WER results of 5-gram and LSTM LMs with different numbers of history utterances.}
    \begin{tabular}{|c|c|c|c|c|c|}
    \hline
    \#hist sent     &  \multicolumn{2}{|c|}{ref} & 
    \multicolumn{2}{|c|}{hyp}     \\
    \cline{2-5}
                    & w/ $<$s$>$ & w/o $<$s$>$ & w/ $<$s$>$ & w/o $<$s$>$\\
    \hline\hline
      0     &   \multicolumn{4}{|c|}{16.20}                 \\
      \hline
      1     &  15.80    &  15.67    &  15.84  &  15.75       \\
      2     &   15.69   &  15.66     & 15.78  &   15.64    \\
      4     &   15.71    &  15.54    & 15.66  &   15.59  \\
      $+\infty$& 15.59   &  15.51     &  15.66  &    15.60       \\
    \hline        
    \end{tabular}
    \label{tab:werlstmlonghist}
\end{table}

As discussed in Section \ref{sec:context}, in previous work, the long-term history information can be incorporated into LSTM-LMs 
by computing LSTM-LMs probability with history in the second pass rescoring. For the same hypothesis, scores
from the first and second pass should be exactly the same. The only difference lies in the search space.
It is worth noting that using $n$-gram LM without history and LSTM-LM with history have different search space. The hypothesis has a high likelihood in the later case is possible to have a relatively low likelihood in the first case.
In order to verify that the WER improvement is indeed from the use of context in the first-pass decoding at early stage,
the last experiment compares WERs for context used in the first-pass decoding and the second-pass rescoring.  The $n$-best ($n$=100) list was generated using the $n$-gram LM and rescored by LSTM-LM containing the complete reference context. The WER results are presented in Table \ref{tab:wernbesthistlm}. It can be seen that when the context is applied for the second pass rescoring, the WER improvement is 0.3\% to 0.4\% worse than the use of history in the first pass decoding. This indicates that the n-best list generated by $n$-gram LMs is not able to fully utilize the context information and the correct paths might be pruned early during beam search.

\begin{table}[htbp]
    \centering
    \caption{WER results of the first-pass and second-pass rescoring for LSTM-LMs with reference and hypothesis history information.}
    \begin{tabular}{|c||c|c|c|}
    \hline
    \#pass       & ref       & hyp \\
    \hline\hline
    $1^{st}$     & 15.59     &  15.66     \\
    $2^{nd}$     & 15.90     &  16.01 \\
    \hline
    \end{tabular}
    \label{tab:wernbesthistlm}
\end{table}

\section{Conclusions and Future Work}
\label{sec:conclusion}
In this paper, the use of context information was investigated for LSTM-LMs based first-pass decoding in a WFST decoder.
The long-term history nature of LSTM-LMs allows the LSTM-LMs trained on utterance level be applied to capture the context information across utterance and guide the search effectively at the early stage.
The experimental results on the internal meeting test set show that LSTM-LMs in first-pass decoding provided comparable performances compared to second-pass rescoring when no context information is used. In contrast, when the long-term history is available, the use of history information for LSTM-LMs in the first-pass decoding yielded significant WER performance improvement. It also outperformed the use of context information in second-pass rescoring for LSTM-LMs.

There are several future directions we are going to explore. The current work adopted LSTM-LMs trained on utterance level and evaluation is operated across utterance. It would be useful to train LSTM-LMs in the paragraph level to allow them more capable of capturing information across utterance. The other direction we want to inspect is how to incorporate the context information more effectively.

\section{Acknowledgements}
\label{ssec:acknowledgement}
The authors would like to thank Hosam Khalil, Veljko Miljanic and Yuhui Wang for the fruitful discussion and help on the WFST decoder.

\newpage
\bibliographystyle{IEEEtran}
\bibliography{refs}

\end{document}